\documentclass{llncs}
\usepackage{times}
\usepackage{balance}
\usepackage{mathptmx}
\usepackage{amsmath,epsfig,hyperref}
\usepackage{graphics,graphicx}
\usepackage{sidecap}
\usepackage{subfigure}
\usepackage{longtable}
\usepackage{eso-pic}
\usepackage{amsmath,graphicx}
\usepackage{times}
\usepackage{balance}
\usepackage{mathptmx}
\usepackage{amsmath,epsfig,hyperref}
\usepackage{graphics,graphicx}
\usepackage{sidecap}
\usepackage{subfigure}
\usepackage{longtable}
\usepackage{eso-pic}
\usepackage{amsmath,epsfig}
\usepackage{graphics,graphicx}

\usepackage{times}
\usepackage{balance}
\usepackage{mathptmx}
\usepackage{amsmath,epsfig,hyperref}
\usepackage{graphics,graphicx}
\usepackage{sidecap}
\usepackage{subfigure}
\usepackage{longtable}

\usepackage{array}
\usepackage{times}
\usepackage{balance}

\usepackage{amsmath,graphicx}
\usepackage{times}
\usepackage{balance}
\usepackage{mathptmx}
\usepackage{amsmath,epsfig,hyperref}
\usepackage{graphics,graphicx}
\usepackage{sidecap}
\usepackage{subfigure}
\usepackage{longtable}
\usepackage{eso-pic}
\usepackage{amsfonts}
\usepackage{amssymb}
\usepackage{multicol}
\usepackage{url}
\urldef{\mailsa}\path|{a.abdelkher,mohamed.elhassouni}@gmail.com|
\urldef{\mailsb}\path|hocine.cherifi@u-bourgogne.fr|
\urldef{\mailsc}\path|aboutaj@fsr.ac.ma|
\usepackage{longtable}
\newcommand{\argmax}{\operatornamewithlimits{arg max}}
\newcommand{\argmin}{\operatornamewithlimits{arg min}}

\everymath{\displaystyle}

\begin{document}

\title{Image quality assessment measure based on natural image statistics in the Tetrolet domain}

%
\author{Abdelkaher Ait Abdelouahad$^1$%
\and Mohammed El Hassouni$^2$%
\and Hocine Cherifi$^3$
\and Driss Aboutajdine$^1$
}
%
\institute{$^1$LRIT URAC- University of Mohammed V-Agdal-Morocco\\
$^2$DESTEC, FLSHR- University of Mohammed V-Agdal-Morocco\\
$^3$Le2i-UMR CNRS 5158 -University of Burgundy, Dijon-France\\
\mailsa\\
\mailsb\\
\mailsc\\}

\maketitle              
\begin{abstract}
This paper deals with a reduced reference (RR) image quality measure based on natural image statistics modeling. For this purpose, Tetrolet transform is used since it provides a convenient way to capture local geometric structures. This transform is applied to both reference and distorted images. Then, Gaussian Scale Mixture (GSM) is proposed to model subbands in order to take account statistical dependencies between tetrolet coefficients.  In order to quantify the visual degradation, a measure based on Kullback Leibler Divergence (KLD) is provided. The proposed measure was tested on the  Cornell VCL A-57 dataset and compared with other measures according to FR-TV1 VQEG framework.
\end{abstract}
\section{Introduction}
Recently, several RR methods have been introduced but few of them are general-purpose. The first general-purpose RR methods was introduced by Wang \cite{ex1} in the steerable pyramids domain named WNISM. The  KLD was used to quantify the difference between two subband coefficient histograms. The first histogram is computed from the distorted image while the second is  summarized using the Generalized Gaussian Density (GGD) model parameters instead of sending all histogram bins. Promising results were obtained for five distortions in the LIVE dataset. Tao et al \cite{ex2} have proposed the contourlet transform which is effective in dealing with directional information like edges. After CSF masking, the JND is applied to remove visually insensitive coefficients. A histogram is formed from the remaining coefficient. Finally, the histogram is normalized and considered as RR feature. Results were presented for two distortions from the LIVE dataset : JPEG and JPEG2000 compressions. Li \emph{et al} \cite{ex3} investigated the Divisive Normalization Transform (DNT) to take into account the dependencies between wavelet coefficients which were ignored in the WNISM. The measure based on the DNT improved the WNISM, specially when it was tested on a set formed by different distortions. Nevertheless, its performances can change significantly since it depends on some parameters which need to be trained. In \cite{ex4} the construction of the Strongest Component Map (SCM) is proposed. The Weibull distribution parameters are estimated from the SCM coefficients histograms. Finally, only the scale parameter $\beta$ is involved in a measure called $\beta$W-SCM. Experiments with the LIVE dataset show significant correlation between the model predictions and the subjective scores, nearly the same as WNISM. In \cite{ex5} grouplets have been used to capture image geometric structures and orientations. To incorporate  HVS characteristics, a Contrast Sensitivity Function (CSF) is applied before measuring the changes between the reference and the distorted images. Their results show some significant improvements for JPEG distorted images as compared to WNISM.\\ Inspired by the work of Wang, we have proposed the use of the BEMD (Bi-dimentional Empirical Mode Decomposition) in the general scheme as an adaptive decomposition. Although the BEMD-based method outperforms the WNISM over several distortions in the TID 2008 dataset, low correlations with human judgment were obtained.\\In this work, we propose a joint probability distribution of tetrolet coefficients using GSM model. This allows us to exploit the dependencies between tetrolet coefficients. A GSM model is defined as the product of zero mean Gaussian vector and positive random variable called multiplier. Here, we propose Weibull distribution to model the multiplier distribution. Then, assuming the independency between GSM components (the multiplier and the Gaussian vector) we derived an expression for the KLD in order to evaluate the visual quality of a processed image.\\The rest of this paper is organized as follows. In section 2 we give a brief review of the tetrolet transform, in section 3 we explain how we model the dependencies between tetrolet coefficients using the GSM model, we present the distortion measure in section 4. Section 5 is reserved for experimental results and finally a conclusion ends the paper.
\section{Tetrolet transform}
Nowadays, a sparse representation is required in image processing techniques. In such representation the energy of the signal is concentrated in few number of coefficients not null. This facilitates the feature extraction step used in image retrieval, image classification and RRIQA algorithms. Although wavelets were introduced for this aim, they can take advantage only of singularity points. Thus directional information like edge is disregarded. The idea of tetrolet transform \cite{ex6} is to allow more general partitions which capture the image local geometry by bringing the "tiling by tetrominoes" problem into play.\\Tetrominoes are derived from the well know game "tetris". They were introduced by Golomb \cite{ex7}. We can obtain a tetromino by connecting four equal sized square.  Disregarding rotation and isometric we have five free tetrominoes as shown in Figure 1.
\begin{figure}[htb]
  \centering
 \centerline{\epsfig{figure=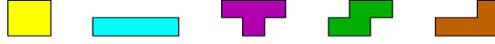,width=10.0cm,height=2.0cm}}
\vspace {-1cm}
\caption{The five free tetrominoes.}
\end{figure}
The Haar transform is a special case, since it considers only the first tetromino (square). To use other tetrominoes we should have at least a $4\times4$ blocks (Figure 2) which will give 117 possibility, whereas a $8\times8$ blocks gives $117^4> 10^8$ possibilities. From computational complexity standpoint, it's clear that the first choice is the reasonable one.
\begin{figure}[htb]
  \centering
 \centerline{\epsfig{figure=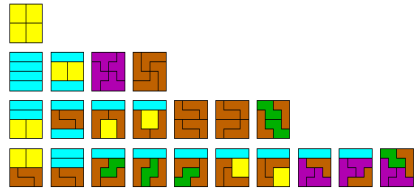,width=7.0cm,height=3.5cm}}
\vspace {-1cm}
\caption{The 22 fundamental forms tiling a $ 4 \times 4$ board.}
\end{figure}
Therefore,  tetrominoes ensure more directions when rotations and reflections are considered. To illustrate this let's take from Figure 2 (Line 4) the third covering (from left to right), eight other coverings are possible with different directions are shown in Figure 3.
\begin{figure}[htb]
  \centering
 \centerline{\epsfig{figure=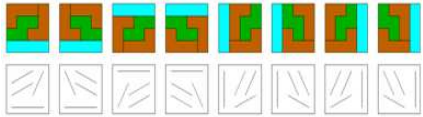,width=8.0cm,height=3.0cm}}
\vspace {-1cm}
\caption{Different directions covered by the same tetrominoes.}
\label{fig:res}
\end{figure}
\subsection{Computing Tetrolet transform}
The computation of the tetrolet transform consists in two stages. First, the tiling by tetrominoes is achieved ensuring an optimal covering for each $4\times4$ block $Q_{i,j}$ in the image. Second, the Haar transform is applied to the tetrominoes of each covering. More precisely, let us take an image $\mathrm{a}^0=[a(i,j)]^{N-1}_{i,j=0}$ , $N$ is a power of $2$ , i.e $N=2^p, p \in \mathbb{N}$  and we suppose that we are in the $r^{th}$ level. The image is decomposed into $4\times4$ blocks, for each block we consider the 117 possible covering $c=1,...,117$. The Haar transform is than applied to the tetrominoes forming the investigated covering. This leads to four low-pass coefficients and 12 tetrolet coefficients as follows :
\begin{eqnarray*}
\mathrm{a}^{r,(c)} =(a^{r,(c)}[s])^3_{s=0} \hspace{1cm} and \hspace{1cm} \mathrm{w}_l^{r,(c)}=(w_l^{r,(c)}[s])^3_{s=0}
\end{eqnarray*}
$c$ and $r$ refer to the actual covering and the actual level of decomposition respectively, while $s$ refers to the tetrominoes of the covering and $l$ refers to the three high-pass parts.\\The optimal covering $C_{op}$ is then qualified as the one whose tetrolet coefficients provide the minimal $l^1$  :
\begin{align}C_{op} & = \argmin_{c}   \sum_{l=1}^{3} ||\mathrm{w}_l^{r,(c)}||_1 \notag\\ \
        &= \argmin_{c} \sum_{l=1}^{3}\sum_{s=0}^{3} |w_l^{r,(c)}[s]| \end{align}
In other words, the smaller is the magnitude of the 12 tetrolet coefficients, the  minimal is the $l^1$ norm. Thus we obtain the optimal covering and a sparse image representation.\\Once we get the optimal covering $C_{op}$, we store the corresponding four low-pass coefficients and 12 tetrolet coefficients : $[\mathrm{a}^{r,(c_{op})}, \mathrm{w}_1^{r,(c_{op})}, \mathrm{w}_2^{r,(c_{op})}, \mathrm{w}_3^{r,(c_{op})}]$. Doing this for all blocks $Q_{i,j}$ in the image we achieve the tetrolet transform. Before applying further levels of the tetrolet transform, we should rearrange the components of the vector $\mathrm{a}^{r,(c_{op})}$ into  $2\times2$ matrix using a reshape function :
\begin{eqnarray}
\mathrm{a}^r_{|Q_{i,j}}=R(\mathrm{a}^{r,(c_{op})})=\begin{pmatrix} a^{r,(c_{op})}[0] & a^{r,(c_{op})}[2] \\ a^{r,(c_{op})}[1] & a^{r,(c_{op})}[3] \end{pmatrix}
\end{eqnarray}
\section{Joint statistics of tetrolet coefficients}
The tetrolet transform provides a multi-resolution representation with three orientations since it is derived from the Haar wavelet transform. Here, we propose to exploited the dependencies between tetrolet coefficients as it was done for wavelet coefficients \cite{ex8} as the same as for the curvelet coefficients \cite{ex9}. The Gaussian Scale mixture (GSM) model has been used to model both marginal and joint statistics of natural image wavelet coefficients \cite{ex10}. Let us consider a N-length random vector $Y$. we assume that $Y$ in our study is formed from coefficients clustered around a given coefficient $y^{s,o}$ at scale $s$ and orientation $o$. $Y$ is a GSM if it can be written as the product of a zero mean Gaussian random vector $U$ with covariance matrix $M$ and a positive scalar random variable $x$ called multiplier:
\begin{eqnarray}
Y \dot{=} x.U
\end{eqnarray}
$\dot{=}$ denotes equality in probability. $U$ and $x$ are independent. If we denote $p_x(x)$ as the density of the variable $x$ the density of $Y$ can be expressed as \cite{ex10}:
\begin{eqnarray}
p_Y(Y)=  \int_{}^{}\frac{1}{[2\pi]^{\frac{N}{2}}|x^2M|^\frac{1}{2}} \exp \bigg(-\frac{Y^TM^{-1}Y}{2x^2}\bigg)p_x(x)dx
\end{eqnarray}
To obtain an explicit expression of the PDF of $Y$ we should specify the density of the multiplier $x$. Since the multiplier variable is positive, several distributions can be considered. Here, we choose  Weibull density. To this end, we should estimate first the multiplier. As this later is unknown, we can estimate it by maximum-likelihood method \cite{ex10}  of the observed coefficients given by :
\begin{align}\hat{x} &= \argmax_{x} \{\log p(Y|x)\} \notag\\ \  &= \argmin_{x} \{N\log x+ \frac{Y^TM^{-1}Y}{2x^2} \}\notag \\ \ &= \sqrt{\frac{Y^TM^{-1}Y}{N}}\end{align}
where $M$ is the covariance matrix of the Gaussian vector estimated from the tetrolet coefficients and $N$ is the length of the vector $Y$. Figure 4 illustrates Weibull fitting to the estimated multiplier.
\begin{figure}[!h]
  \centering
 \centerline{\epsfig{figure=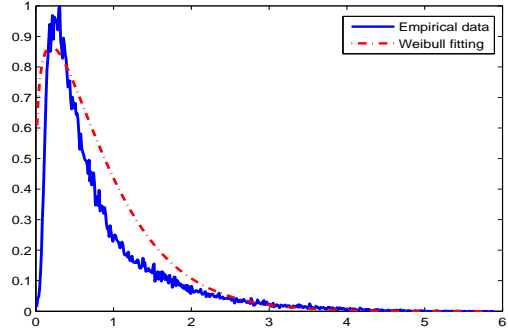,width=8.0cm,height=5.0cm}}
\vspace {-0.5cm}
\caption{Weibull distribution fitted to empirical histogram of the estimated multiplier.}
\label{fig:res}
\end{figure}
\newline The PDF of Weibull distribution is given by:
\begin{eqnarray}
f(x;k,\lambda)= \frac{k}{\lambda}\bigg(\frac{x}{\lambda}\bigg)^{k-1}e^{-(x/\lambda)^k}
\end{eqnarray}
where $k > 0 $ is the shape parameter and $\lambda > 0$  is the scale parameter of the distribution. Inserting the equation (6) in equation (4) the PDF of $Y$ becomes :
\begin{eqnarray}
p_Y(Y)=  \int_{}^{}\frac{kx^{k-1}}{[2\pi]^{\frac{N}{2}}|x^2M|^\frac{1}{2}\lambda^k} \exp \bigg(-\bigg(\frac{Y^TM^{-1}Y}{2x^2}+\bigg(\frac{x}{\lambda}\bigg)^k\bigg)\bigg)dx
\end{eqnarray}
\section{Distortion measure}
In the previous section we have represented the joint statistics of tetrolet coefficients using a univariate Weibull distribution and a multivariate Gaussian distribution. Considering a neighborhood of dimension equals to 9 (3$\times$3).
At the sender side, we apply two levels tetrolet transform to the reference image. This leads to six tetrolet coefficients subbands (2 scales $\times$ 3 orientations). From each subband three features are extracted : the covariance matrix $M$ and the Weibull parameters $(\lambda, k)$. The extracted features are considered as RR side information. Similarly, the same features are extracted from the distorted image at the receiver side and we consider them as reduced description (RD).\\A dissimilarity measure is required to compare the RR to the RD and thus quantify the visual degradation.  According to our knowledge a closed analytical form of the KLD for the proposed joint distribution in equation (7) does not exist. To resolve this problem, let us consider two joint distributions $P_1(Y;M_1,k_1,\lambda_1)$ and $P_2(Y;M_2,k_2,\lambda_2)$, where $Y$ is a GSM vector. Since the components of the GSM (the multiplier and the Gaussian vector) are independent, we can derive an expression for the KLD between two joint distributions as the sum of the KLD between two multivariate Gaussian densities and the KLD between two Weibull distributions. In other words :
\begin{align}
KLD(P_1(Y;M_1,k_1,\lambda_1)||P_2(Y;M_2,k_2,\lambda_2))&=KLD(P_1(x;k_1,\lambda_1)||P_2(x;k_2,\lambda_2))\notag\\&+KLD(P_1(U;M_1)||P_2(U;M_2))
\end{align}
Now, that we have a closed analytical form for the KLD for both, Weibull distribution and the multivariate Gaussian density we can easily derive the KLD for the proposed joint distribution as:
\begin{align}
KLD(P_1(Y;M_1,k_1,\lambda_1)||P_2(Y;M_2,k_2,\lambda_2))&=\Gamma\bigg(\frac{\lambda_2}{\lambda_1}+1\bigg)\bigg(\frac{k_1}{k_2}\bigg)^{\lambda_2} + \ln(k_1^{-\lambda_1}\lambda_1) - \ln(k_2^{-\lambda_2}\lambda_2)\notag\\&+\ln (k_1)(\lambda_1-\lambda_2) +
\gamma\frac{\lambda_2}{\lambda_1}-\gamma-1 \notag\\&+0.5 \bigg[tr(M_2^{-1}M_1) + ln \bigg(\frac{|M_2|}{|M_1|}\bigg)-N\bigg]
\end{align}
where $\gamma$ denotes the Euler-Mascheroni constant ($\gamma\approx0.57721$) and $\Gamma(.)$ is the Gamma function.\\First, the distance in equation (9) is computed to quantify the dissimilarity between two tetrolet coefficient subbands, the first from the reference image and the second is its correspondent from the distorted image. Finally, the dissimilarities between the subbands are combined to produce a global dissimilarity as follows :
\begin{eqnarray}
Q=\log_2(1+\frac{1}{D_0}\sum_{i=1}^{L}D_i)
\end{eqnarray}
where $L$ is the number of the subbands, $D_0$ is a constant to control the scale of the distortion measure and it is equal to 0.1.The log function is involved here to reduce the difference between a high values and a low values of $D$, so that we can have values in the same order.
\section{Experimental results}
Our experimental test was carried out using the Cornell VCL-A 57 \cite{ex11} dataset. It provides 60 distorted images. Three reference images  are altered with six distortions labeled : FLT, NOZ, JPG, JP2, DCQ and BLR. The labels refer to quantization of the LH subbands of a five-level DWT of the image using the 9/7 filters, additive Gaussian white noise, baseline JPEG compression, JPEG2000 compression using the 9/7 filters, JPEG2000 compression using the 9/7 filters with the dynamic contrast-based quantization algorithm, blurring by using a Gaussian filter, respectively. Each image in the Cornell VCL-A57 has its Mean Opinion Score (MOS). The subjective scores must be compared in term of correlation with the objective scores. These objective scores are computed from the values generated by the objective measure, using  a non linear function according to the Video Quality Expert Group (VQEG) Phase I FR-TV \cite{ex12}. Here, we use a four parameters logistic function.\\
\begin{equation}logistic(\gamma,Q)=\frac{\gamma_1-\gamma_2}{1+e^-(\frac{D-\gamma_3}{\gamma_4})}+\gamma_2 \end{equation} where $\gamma=(\gamma_1,\gamma_2,\gamma_3,\gamma_4)$.\\
 Thus, the predicted MOS is given by :
  \begin{equation}
  MOS_p=logistic(\gamma,Q)
  \end{equation}
Once the nonlinear mapping is achieved, we obtain the predicted objective quality scores. To compare the subjective and objective quality scores, several metrics were introduced by the VQEG. In our study, we compute the correlation coefficient to evaluate the accuracy prediction and  the Rank order coefficient to evaluate the monotonicity prediction. Table 1 shows the results for the Cornell VCL A-57 dataset.\\
\begin{table}[!h]
 \caption{Performance evaluation for the  proposed measure using Cornell VCL A-57 dataset.}
  \centering
 \begin{tabular}{|> {\centering} p{2cm} |> {\centering} p{1cm}| > {\centering} p{1cm}| > {\centering} p{1cm}| > {\centering} p{1cm}|> {\centering} p{1cm}|> {\centering} p{1cm}|> {\centering} p{1cm}|}					
\hline
   Dataset &FLT&JPG&JP2&DCQ& BLR& NOZ& All \tabularnewline\hline
   \multicolumn{8}{c}{Correlation Coefficient}  \tabularnewline\hline
        $Proposed$ &0.71& 0.96&0.83&0.95&0.91 & 0.86&0.70\tabularnewline\hline	
        $DNT$ &0.76&0.91 &0.81 &0.90 &0.93 &0.99 &0.66  \tabularnewline\hline
         Method in \cite{ex13} &0.49& 0.85& 0.78& 0.93& 0.76& 0.62& 0.31\tabularnewline\hline
        $PSNR$ &0.91& 0.70& 0.79& 0.56& 0.59& 0.93& 0.63\tabularnewline\hline	
        $MSSIM$ &0.92& 0.91& 0.87& 0.94& 0.79& 0.88& 0.72\tabularnewline\hline	
 \multicolumn{8}{c}{Rank-Order Correlation Coefficient }  \tabularnewline\hline
    $Proposed$ &0.46& 0.96&0.81 &0.90 &0.90 & 0.80&0.74\tabularnewline\hline	
    $DNT$ & 0.50 &0.76 &0.80 &0.66 &0.80 &0.98 &0.70  \tabularnewline\hline	
    Method in \cite{ex13} &0.10& 0.76& 0.53& 0.80& 0.66& 0.73& 0.29\tabularnewline\hline
   $PSNR$ &0.90& 0.63& 0.80& 0.50& 0.46& 0.95& 0.62\tabularnewline\hline	
   $MSSIM$ &0.96& 0.93& 0.86& 0.96& 0.90& 0.91& 0.78\tabularnewline\hline	
 \end{tabular}
\end{table}
As we can see, results reported in table. 1 concern the proposed measure as well as some FR and RR methods. In comparison with RR methods, the proposed measure outperforms the DNT-based methods for JPG, JP2 and DCQ distortions, and the method in \cite{ex13} for JPG, JP2, DCQ, BLR and NOZ distortions. The proposed measure outperforms also the PSNR for JPG, JP2, DCQ and BLR distortions, and MSSIM \cite{ex14}for JPG, DCQ and BLR distortions. However, the proposed measure fails for the FLT distortion.
\section{Conclusion}
In this paper we have introduced a RR measure in the tetrolet domain. The GSM model was used to characterize the dependencies between tetrolet coefficients. We have proposed the Weibull distribution to model the multiplier of the GSM model, this leads to  a new joint distribution. Assuming the independence between GSM components we have derived a closed expression of the KLD for the propose joint distribution. Significant improvements were remarked for the proposed measure when it was tested on the Cornell VCL-A57 dataset.


\begin{thebibliography}{}
%
\bibitem{ex1}
 Z. Wang and E.P. Simoncelli, "Reduced-reference image quality assessment using a wavelet-domain natural image statistic
model," \emph{in Proc.of SPIE Human Visionand Electronic Imaging}, 2005, vol. 5666, pp. 149–-159.
\bibitem{ex2}
 D. Tao, X. Li, W. Lu, and X. Gao, "Reduced-reference IQA in contourlet domain," \emph{IEEE Transactions on Systems, Man,
and Cybernetics, PartB: Cybernetics}, vol. 39, no. 6, pp. 1623–-1627,2009.
\bibitem{ex3}
 Q. Li and Z. Wang, "Reduced-reference image quality assessment using divisive normalization-based image representation,"
\emph{IEEE Journal of Selected Topicsin Signal Processing}, vol. 3, no. 2, pp. 202–-211,2009.
\bibitem{ex4}
 W. Xue and X. Mou, "Reduced reference image quality assessment based on weibull statistics," \emph{in The International
Workshop on Quality of Multimedia Experience (QoMEx)}, 2010, pp. 1–-6.
\bibitem{ex5}
 A. Maalouf, M.C. Larabi, and C. Fernandez-Maloigne, "A grouplet-based reduced reference image quality assessment,"
\emph{in The International Workshop on Quality of Multimedia Experience(QoMEx)}, 2009, pp. 59–-63.
\bibitem{ex6}
 J. Krommweh, "Tetrolet transform: A new adaptive haar wavelet algorithm for sparse image representation," \emph{Journal of
Visual Communication and Image Representation}, vol.21, no. 4, pp. 364–-374, 2010.
\bibitem{ex7}
 S.W.Golomb, Polyominoes: puzzles, patterns, problems,and packings, Princeton Univ Pr,1996.
\bibitem{ex8}
 E.P. Simoncelli, "Modeling the joint statistics of images in the wavelet domain," \emph{in Proc. SPIE}, 1999, vol. 3813, pp.
188–-195.
\bibitem{ex9}
L. Boubchir and J.M. Fadili, "Multivariate statistical modeling of images with the curvelet transform," \emph{in Proc. IEEE
Conf.on Signal Processing and Its Applications}, 2005, pp. 747–-750.
\bibitem{ex10}
 M.J. Wainwright and E.P. Simoncelli, "Scale mixtures of gaussians and the statistics of natural images," \emph{Advances in
neural information processing systems}, vol. 12, no. 1, pp. 855–-861, 2000.
\bibitem{ex11}
 D. M. Chandler and S. S. Hemami, "Cornell-vcl a57 database," Available at:
\emph{http://foulard.ece.cornell.edu/dmc27/vsnr/vsnr.html.,2007.}
\bibitem{ex12}
 A.M.Rohaly, J.Libert, P.Corriveau, A.Webster et al, "Final report from the video quality experts group on the validation
of objective models of video quality assessment," \emph{ITU-T Standards Contribution COM}, pp. 9–-80.
\bibitem{ex13}
 Z. Wang, G. Wu, H.R. Sheikh, E.P. Simoncelli, E.H. Yang, and A.C. Bovik, "Quality-aware images," \emph{IEEE Transactions
on Image Processing}, vol. 15, no. 6, pp. 1680–-1689, 2006.
\bibitem{ex14}
 Z. Wang, A.C. Bovik, H.R. Sheikh, and E.P. Simoncelli, "Image quality assessment: From error visibility to structural
similarity," \emph{IEEE Transactions on Image Processing}, vol. 13, no. 4, pp. 600–-612, 2004.
\end{thebibliography}
\end{document}